\pdfoutput=1

\documentclass[11pt]{article}

\usepackage[]{acl}

\usepackage{times}
\usepackage{latexsym}

\usepackage[T1]{fontenc}

\usepackage[utf8]{inputenc}

\usepackage{microtype}
\usepackage{graphicx}

\usepackage{amsmath}
\usepackage{amsfonts}
\usepackage{multicol}  
\usepackage{multirow}
\usepackage{booktabs}
\usepackage{colortbl}
\usepackage{xspace}

\usepackage{arydshln}

\usepackage{float}

\usepackage{arydshln}
\usepackage{tikz}
\usepackage{pgfplots}

\usepackage{pifont}
\newcommand{\cmark}{\ding{51}}
\newcommand{\xmark}{\ding{55}}

\newcommand{\spider}{\textsc{Spider}\xspace}
\newcommand{\syn}{\textsc{Syn}\xspace}
\newcommand{\dk}{\textsc{Dk}\xspace}
\newcommand{\real}{\textsc{Realistic}\xspace}
\newcommand{\squall}{\textsc{Squall}\xspace}

\newcommand{\picard}{\textsc{Picard}\xspace}
\newcommand{\sun}{\textsc{Sun}\xspace}

\newcommand{\sqlcorrect}[1]{{\color{olive}{\cmark}}}
\newcommand{\sqlwrong}[1]{{\color{red}{\xmark}}}

\def\eg{\emph{e.g.}}

\makeatletter
\renewcommand{\maketag@@@}[1]{\hbox{\m@th\normalsize\normalfont#1}}
\makeatother
\title{\sun: Exploring Intrinsic Uncertainties in Text-to-SQL Parsers}

\author{Bowen Qin$^{1, 2, \includegraphics[scale=0.02]{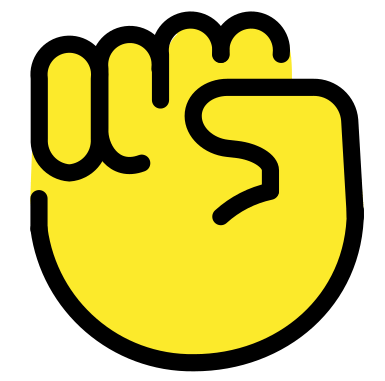}}$, Lihan Wang$^{1, 2, \includegraphics[scale=0.02]{first.png}}$, Binyuan Hui$^{3, \includegraphics[scale=0.02]{first.png}}$, Bowen Li$^{3}$, Xiangpeng Wei$^{3}$ \\
\bf{Binhua Li$^{3}$, Fei Huang$^{3}$, Luo Si$^{3}$, Min Yang$^{1}$\footnotemark[2], Yongbin Li$^{3}$\footnotemark[2]}\\
$^1$ Shenzhen Institute of Advanced Technology, Chinese Academy of Sciences\\
$^2$ University of Chinese Academy of Sciences \\
$^3$ DAMO Academy, Alibaba Group \\
\texttt{\{bw.qin, lh.wang1, min.yang\}@siat.ac.cn} \\
\texttt{\{binyuan.hby, xiangpeng.wxp, binhua.lbh, shuide.lyb\}@alibaba-inc.com}\\
}

\begin{document}
\maketitle
\renewcommand{\thefootnote}{\fnsymbol{footnote}}
\footnotetext{\includegraphics[scale=0.02]{first.png} Equal contribution.}
\footnotetext[2]{Corresponding authors.}
\begin{abstract}
This paper aims to improve the performance of text-to-SQL parsing by exploring the intrin\textbf{\underline{s}}ic \textbf{\underline{un}}certainties in the neural network based approaches (called \textbf{\sun}). From the data uncertainty perspective, it is indisputable that a single SQL can be learned from multiple semantically-equivalent questions.
Different from previous methods that are limited to one-to-one mapping, we propose a data uncertainty constraint to explore the underlying complementary semantic information among multiple semantically-equivalent questions (many-to-one) and learn the robust feature representations with reduced spurious associations. 
In this way, we can reduce the sensitivity of the learned representations and improve the robustness of the parser.
From the model uncertainty perspective, there is often structural information (dependence) among the weights of neural networks. 
To improve the generalizability and stability of neural text-to-SQL parsers, we propose a model uncertainty constraint to refine the query representations by enforcing the output representations of different perturbed encoding networks to be consistent with each other. 
Extensive experiments on five benchmark datasets demonstrate that our method significantly outperforms strong competitors and achieves new state-of-the-art results.
For reproducibility, we release our code and data at \url{https://github.com/AlibabaResearch/DAMO-ConvAI/tree/main/sunsql}.
\end{abstract}

\section{Introduction}
Text-to-SQL parsing \cite{zettlemoyer2012learning, liang2013learning, zhong2017seq2sql, Qin2022ASO} aims at converting a natural language  (NL) question to its corresponding structured query language (SQL) in the context of a relational database (Schema).
Although relational databases can be efficiently accessed by skilled professionals via handcrafted SQLs, a natural language interface, whose core component relies on text-to-SQL parsing, would allow ubiquitous relational data to be accessible to a broader range of non-technical users. Therefore, text-to-SQL parsing has attracted increasing attention from both academic and industrial communities recently due to its broad applications in question answering, conversational search interaction, and so on.

Although significant efforts have been devoted to text-to-SQL parsing \cite{chen2021shadowgnn, typesql, ratsql, lgesql, Scholak2021PICARDPI} with advanced deep models and architectures by learning black-box mappings between input NL questions and output SQLs, there are still several technical challenges for accurate and robust text-to-SQL parsing.
First, previous models are generally learned to fit the simplified one-to-one mapping relationship, where only one NL question is used as the input, and the appropriate rest are ignored. 
However, there exists data uncertainty \cite{ott2018analyzing,wei2020uncertainty} in text-to-SQL parsing, i.e., one output SQL may correspond to multiple semantically-equivalent NL questions. 
At inference time, the semantic parser trained on the one-to-one parallel data struggles to deal with adequate variations of the training queries. 
Second, there is often structural information (dependence)  \cite{xiao2019quantifying, zhang2022struct} among the weights of neural networks. 
One challenge in training neural semantic parsers is that such models may overfit the training data since these models only seek a point estimate for their weights, failing to quantify weight (model) uncertainty. 
For text-to-SQL parsing, the model uncertainty brings difficulty in obtaining the encoded representations that can best describe the input data distribution and provide an robust mapping between NL questions and SQL queries.

To alleviate the aforementioned challenges, in this paper, we propose a generic training approach \sun, which explores the data and model uncertainties in text-to-SQL parsing. 
First, we propose a data uncertainty constraint to explore the underlying complementary semantic information among multiple semantically-equivalent queries and learn comprehensive feature representations with strong expressive ability. 
In particular, we summarize multiple semantically-equivalent source questions into a closed semantic region which is then used to complement the model to generate better SQL queries with comprehensive semantics.
Second, to improve the generalizability and stability of neural text-to-SQL parsers, we propose a model uncertainty constraint to refine the query representations by enforcing the output representations of different perturbed encoding networks to be consistent with each other. 
Concretely, we impose the consistency on multiple networks perturbed with dropout for the same input NL question. 

We summarize our main contributions as follows. (1) We propose a data uncertainty constraint, which aims to explicitly capture comprehensive semantic information among multiple semantically-equivalent NL questions, and enhance the hidden representations with this complementary information for generating better SQL, thus improving the robustness of neural text-to-SQL parsers. (2) We employ a model uncertainty constraint to encourage high similarity between the output representations of two perturbed encoding networks for the same input NL question, improving the generalizability and stability of neural text-to-SQL parsers. (3) Experiments on five benchmark datasets demonstrate that the proposed \sun method outperforms the strong competitors by a substantial margin. \textit{\textbf{It is noteworthy that our method is model-agnostic and potentially applicable for any text-to-SQL parsers with deep network architectures.}}

\section{Related Work}
\paragraph{Text-to-SQL Parsing}
Text-to-SQL parsing, a subtask of semantic parsing, aims at converting a NL question to its corresponding SQL query in the context of a relational database (schema). 
Inspired by the success of deep learning, neural text-to-SQL models based on the sequence-to-sequence (Seq2Seq) framework have dominated the research field of text-to-SQL parsing \cite{guo2019towards,wang2020relational,zhong2020grounded,hui2021dynamic,hui2021improving,hui2022ssql,lgesql, wang2022proton}. 
The general idea behind these methods is to construct an encoder to encode the input question together with related table schema and leverage a decoder to generate the target SQL based on the output of the encoder. 
For example, IRNet \cite{guo2019towards} is a representative neural text-to-SQL parser, which leveraged two separate BiLSTMs with self-attention mechanism \cite{vaswani2017attention} to encode the NL question and table schema. 
Subsequently, the graph-based approaches have been proposed, which use relational graph attention networks to deal with the schema entities and question words with structured reasoning. For instance, RATSQL \cite{ratsql},  SMBOP \cite{rubin2020smbop} and RaSaP \cite{huang2021relation} defined a question-schema graph and employed the relation-aware self-attention mechanism \cite{shaw2018self} in the encoding process to jointly learn representations of question words, schema items and edge relations. LGESQL \cite{lgesql} further constructed an edge-centric graph to update the edge features and designed graph pruning to determine the golden schema items related to the NL question.

Recently, some methods have leveraged the powerful pre-training capabilities of T5 \cite{raffel2019exploring} to generate SQL queries. Different from graph-based methods, T5-based approaches \cite{Scholak2021PICARDPI} adopt the transformer-based architecture for both encoder and decoder and do not need pre-defined graphs, schema linking relations, and grammar-based decoder.
PICARD \cite{Scholak2021PICARDPI} is a representative T5-based method which constrained auto-regressive decoders of language models through incremental parsing. The impressive experimental results verify the ability of the T5-based methods for text-to-SQL parsing.

\paragraph{Uncertainty Modeling in NLP}
Uncertainty quantification is an important approach to building robust AI systems. In the field of natural language processing, there are several works \cite{kendall2015bayesian,xiao2019quantifying, zhang2019mitigating, shen2019modelling, wei2020uncertainty, zhang2021bayesian, hu2021uncertainty} which investigate the effects of quantifying uncertainties in various NLP tasks. For example,  \citet{zhang2019mitigating} applied a dropout-entropy method to measure uncertainty learning for text classification. 
\citet{xiao2019quantifying} showed that explicitly modeling uncertainties via Monte-Carlo dropout \cite{gal2016dropout} could enhance model performances of several NLP tasks.
\citet{su2018variational} introduced a series of continuous latent variables to model underlying semantics of source sentences in neural machine translation. \citet{wang2019improving} proposed to quantify the confidence of NMT model predictions based on model uncertainty to better cope with noise in synthetic corpora.
\citet{wei2020uncertainty} considered the intrinsic uncertainty by representing multiple source sentences into a closed semantic region. 
To our best knowledge, our proposed method is the first attempt to explore uncertainty in text-to-SQL parsing.

\section{Preliminaries}

\subsection{Problem Definition}
Given a natural language question $Q$ and the corresponding database schema $S=\langle T, C\rangle$, text-to-SQL parsing aims to generate a SQL query $Y$ based on $Q$ and $S$.
More specifically, the question $Q = \left\{q_{1}, q_{2}, \cdots, q_{|Q|} \right\}$ is a sequence of tokens, and the schema $S$ consists of tables $T=\left\{t_{1}, t_{2}, \cdots, t_{|\mathcal{T}|} \right\}$ and columns $C=\left\{c_{1}, c_{2}, \cdots, c_{|C|} \right\}$.
Each table $t_i$ contains $m_t$ words $(t_{i,1}, t_{i,2}, \cdots, t_{i,m_t})$ and each column name $c_j^{t_i}$ in table $t_i$ contains $m_c$ words $(c_{j,1}^{t_{i}}, c_{j,2}^{t_{i}}, \cdots, c_{j,m_c}^{t_{i}})$.
We use $I = \langle Q, T, C\rangle$ to denote an input for the text-to-SQL parser.

\subsection{Text-to-SQL parser}
Currently, advanced text-to-SQL parsers are centred around two types of approaches: the graph-based approaches \cite{ratsql,huang2021relation,lgesql} and the T5-based approaches \cite{Scholak2021PICARDPI,UnifiedSKG}, both of which adopt the encoder-decoder framework for implementation.

\paragraph{Graph-based Methods}
Formally, the input question and database schema are constructed as a single direct graph in the pre-processing phase: $G = \langle V, E \rangle $, where $V = Q \cup T \cup C$ denotes the node set that contains three different node types (question, table, and column) and $E$ is the edge set depicting  \emph{pre-existing} relations for question tokens and schema items. 
To obtain the initial representation for every node in the graph, the recent graph-based methods, \eg, LGESQL \cite{lgesql}, first flatten all question words and schema items into a sequence and feed the sequence $I = \langle Q, T, C\rangle$ into large-scale pre-trained language models (PLMs) to learn  word vectors. The learned word vectors are then passed into a sub-word attentive pooling layer and three Bi-LSTMs according to the node types to get the node representations for the graph $G$. 

After that, the graph-based approaches adopt an encoder, which consists of a stack of relational graph attention network (RGAT) \cite{wang2020relational} layers, to learn complex interaction over schema items as well as question words and output the final contextual representation $X_{I}$ for input $I$.

In the decoding process, the graph-based methods usually adopt the grammar-based syntactic neural decoder to generate the abstract syntax tree (AST) of the target query $Y$ in the depth-first traversal order. The output at each decoding timestep is either (i) an \texttt{APPLYRULE} action that expands the current non-terminal node in the partially generated AST, or (ii) an \texttt{SELECTTABLE} or \texttt{SELECTCOLUMN} action that chooses certain schema item. 
The readers can refer to \cite{ratsql} for more implementation details.

\paragraph{T5-based Methods}
The T5-based approaches leverage the powerful pre-training capabilities of T5 to generate SQL queries, which directly fine-tune the downstream corpora with the standard objective for text generation without any specifically designed modules.
Specifically, they take the sequential concatenation of the question words and schema item names as input to the T5 encoder and generate the corresponding SQL using the T5 decoder. Both encoder and decoder are composed of multi-layer Transformer blocks.

To ensure that the output SQL is grammatically correct, T5-based approaches such as \picard \cite{Scholak2021PICARDPI} implement rule-based constrained decoding, achieving competitive performance with less invasiveness and better compatibility. 
\picard is an incremental parsing method for constrained decoding and can be compatible with any existing auto-regressive language model decoder and vocabulary—including, but not limited to, those of large pre-trained transformers.
Different from the graph-based approaches that generally restrict the auto-regressive decoding process to tokens that can correctly parse to abstract syntax trees during training, constrained decoding used in T5-based approaches operates directly on the output of the language model and applies the characteristic of target SQL to help reject inadmissible tokens at each decoding step at inference time.

\begin{figure*}[t]
\centering
\includegraphics[width=0.85\textwidth]{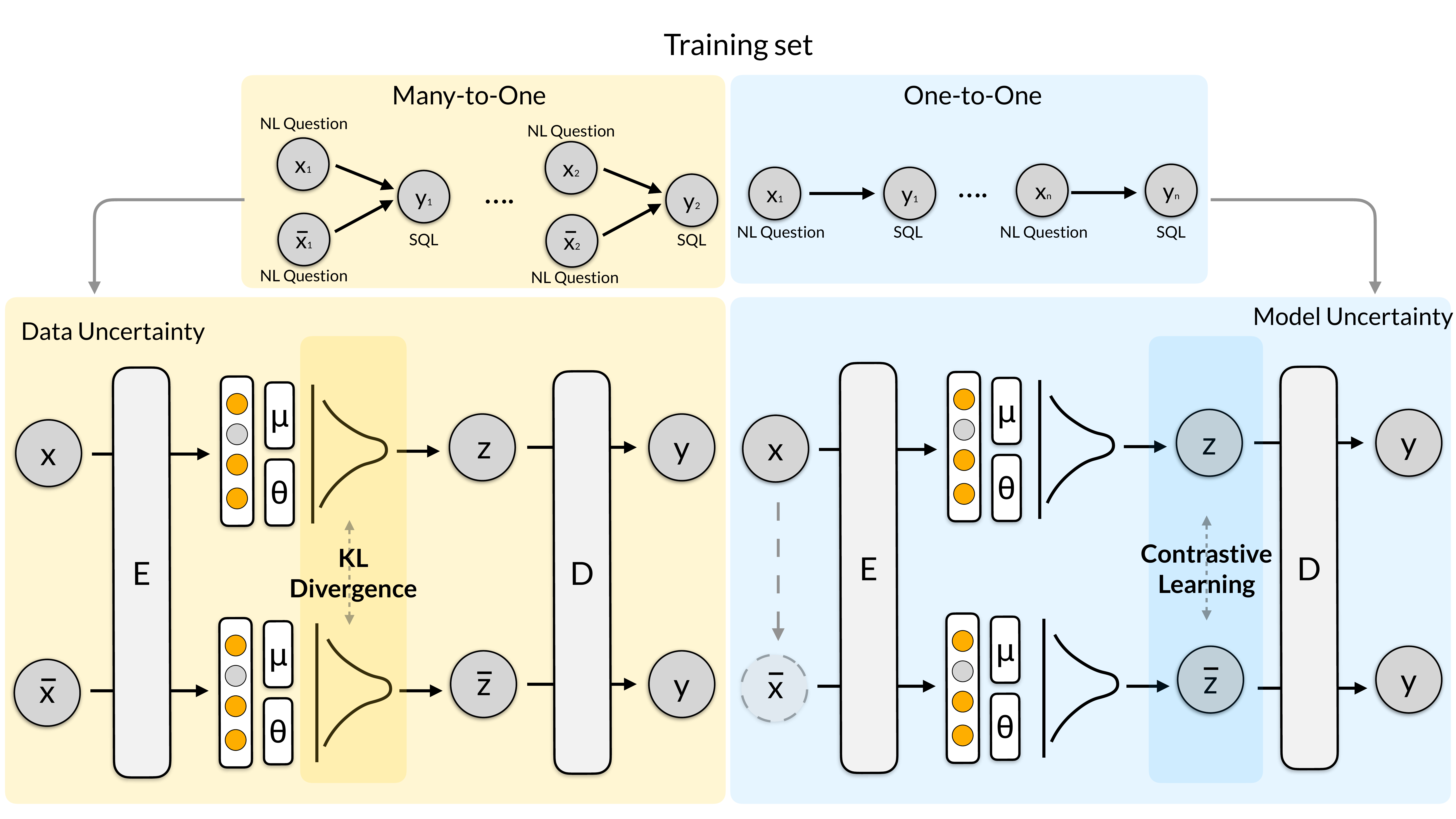}
\caption{Overview of the proposed \sun method. There are many-to-one and one-to-one cases in the text-to-SQL training data. For many-to-one samples, they are modelled by data uncertainty, and for one-to-one samples, they are modelled by model uncertainty. The E means the encoder and D means the decoder of the graph-based or T5-based methods.}
\label{fig:graph}
\end{figure*}

\section{Proposed Method}
\paragraph{Intrinsic Uncertainty} Previous works have focused on the inductive bias through model design, while the intrinsic data uncertainty and model uncertainty are underexplored. 
As shown in Figure \ref{fig:graph}, \textbf{many-to-one} cases (multiple NL questions can correspond to one SQL query) exist in the text-to-SQL training data, which are caused by the inherent uncertainty of natural language \cite{ott2018analyzing, wei2020uncertainty}. 
In such \textbf{many-to-one} scenario, each input sample $I$ has a semantically-equivalent sample $\overline{I}$ in the training set, where these two input instances correspond to the same SQL query.  
In addition, there are also some instances in the training set that do not have semantically-equivalent samples (referred to \textbf{one-to-one} cases), where the model uncertainty becomes critical \cite{xiao2019quantifying,zhang2019mitigating}.
In this work, we propose \sun, a generic training approach for text-to-SQL parsing, to mitigate the data uncertainty and model uncertainty simultaneously.  \sun consists of two primary components: (i) a data uncertainty constraint that summarizes multiple semantically-equivalent NL questions into an abstract semantic region and (ii) a model uncertainty constraint that encourages high similarity between the output representations of two perturbed encoding networks for the same NL question.

\paragraph{Global Semantic Representation Learning}
Without the loss of generality, we use the encoder of the text-to-SQL parser to transform each input $I$ into a contextual representation $X_{I}$. 
Then, we pass the contextual representation $X_{I}$ to a convolutional layer with a pooling operation to obtain the final global semantic representation $H_{I}$ of the input $I$. For Simplicity, we represent the whole process of learning global 
semantic representations as $H_{I}=\texttt{Encode}(I)$.

\subsection{Data Uncertainty Constraint}

For the many-to-one scenario, we expect to encourage high similarity between the representations of two semantically-equivalent NL questions. To this end, we propose a data uncertainty constraint to summarize the semantically-equivalent input $I$ and $\overline{I}$ into a closed semantic region. We first obtain global semantic representations $H_{I}=\texttt{Encode}(I)$ and $H_{\overline{I}}=\texttt{Encode}(\overline{I})$ for two semantically-equivalent input samples $I$ and $\overline{I}$. To learn compact and abstract representation without spurious association, we assume that there are two latent semantic variables $\mathbf{z}$ and $\overline{\mathbf{z}}$ associated with $H_{I}$ and $H_{\overline{I}}$ respectively, which are sufficient for generating the target SQL query and  eliminate the redundant details, so as to reduce the sensitivity of the learned representations. That is, 
both $\mathbf{z}$ and $\overline{\mathbf{z}}$ maintains all information which is shared by $H_{I}$ and $H_{\overline{I}}$, inspired by the intuition that two semantically-equivalent samples provide the same predictive information.  
Formally, we assume that  $\mathcal{P}(\mathbf{z}| H_{I})$ and $\mathcal{P}(\overline{\mathbf{z}}| H_{\overline{I}})$ are the distributions of the latent semantic variables $\mathbf{z}$ and $\overline{\mathbf{z}}$, which are modeled by Normal distributions parametrized with ($\mu$,$\bar{\mu}$) and ($\sigma$, $\bar{\sigma}$). Formally, we define the two semantic distributions as follows:
\begin{align}
&\mathcal{P}(\mathbf{z} \mid H_{I}) \sim \mathcal{N} (\mu(H_{I}), \sigma^{2}(H_{I}) \mathbf{I} ) \\
&\mathcal{P}(\overline{\mathbf{z}} \mid H_{\overline{I}}) \sim \mathcal{N}\left(\bar{\mu}(H_{\overline{I}}), \bar{\sigma}^{2}(H_{\overline{I}}) \mathbf{I}\right) 
\end{align}
where $\mathbf{I}$ denotes the all-ones vector. $\mu$ (or $\bar{\mu}$) and $\sigma$ (or $\bar{\sigma}$) are computed via fully-connected neural networks based on the global semantic representations $H_{I}$ (or $H_{\overline{I}}$) as:
\begin{align}
    \mu(H_{I}) =H_{I} \cdot \mathbf{W}_{\mu}+\mathbf{b}_{\mu} \\
    \log \sigma^{2}(H_{I}) =H_{I} \cdot \mathbf{W}_{\sigma}+\mathbf{b}_{\sigma} \\
    \bar{\mu}(H_{\overline{I}}) =H_{\overline{I}} \cdot \mathbf{W}_{\mu}+\mathbf{b}_{\mu} \\
    \log \bar{\sigma}^{2}(H_{\overline{I}}) =H_{\overline{I}} \cdot \mathbf{W}_{\sigma}+\mathbf{b}_{\sigma}
\end{align}
where $\mathbf{W}_{\mu}$, $\mathbf{W}_{\sigma}$ denote the projection parameters and $\mathbf{b}_{\mu}$, $\mathbf{b}_{\sigma}$ denote the bias terms.

Inspired by the reparameterization techniques used in \cite{kingma2014semi,zhang-etal-2016-variational-neural}, the sampled latent representations $z$ and $\overline{z}$ for $\mathbf{z}$ and $\overline{\mathbf{z}}$ are then obtained as follows for efficient gradient computation :
\begin{align}
z=\mu(H_{I})+\sigma(H_{I}) \odot \epsilon \\
\overline{z}=\bar{\mu}(H_{\overline{I}})+\bar{\sigma}(H_{\overline{I}}) \odot \epsilon
\end{align}
where $\epsilon \sim \mathcal{N}(0, \mathbf{I})$ is a parameter to introduce noise and $\odot$ denotes an element-wise product.

The goal of the data uncertainty constraint is to summarize multiple source questions that share the same meaning into a closed semantic region. 
\textit{In order to enforce the distribution $\mathcal{P}(\mathbf{z}|H_{I})$ to be close to $\mathcal{P}(\overline{\mathbf{z}}|H_{\overline{I}})$, we formulate the data uncertainty constraint loss $\mathcal{L}_{\mathrm{\textbf{DU}}}$ by minimizing KL divergence between $\mathcal{P}(\mathbf{z}|H_{I})$ and $\mathcal{P}(\overline{\mathbf{z}}|H_{\overline{I}})$ as}:
\begin{equation}
{\mathcal{L}_\mathrm{\textbf{DU}}}=\mathbf{K L}\left(\mathcal{P}(\overline{\mathbf{z}} \mid H_{\overline{I}})|| \mathcal{P}(\mathbf{z} \mid H_{I})\right)
\end{equation}


\subsection{Model Uncertainty Constraint}
To improve the robustness of text-to-SQL parsers, we consider the model uncertainty for the one-to-one questions which do not have semantically-equivalent questions in training set by taking advantage of the data augmentation technique. 
Specifically, the input data $I$ goes through the forward pass of the encoding network twice with dropout to produce two-view representations of the input $I$. Since the dropout \cite{hinton2012improving}  operator randomly drops units from the model, the two forward passes indeed produce two distinct semantic representations of input $I$. 
Inspired by the model uncertainty learning in text feature space \cite{zhang2019mitigating}, 
we then use contrastive learning  \cite{gao2021simcse, yan2021consert} to pull together the two-view representations of the same input question produced by dropout and push apart the semantic representations of different questions in the same batch.
\textit{Formally, we formulate the model uncertainty constraint loss $\mathcal{L}_{\rm MU}$ as}:
\begin{equation}
\label{equation 2}
\small
{\mathcal{L}_{\rm MU}} =-\frac{1}{N} \sum_{i=1}^{N} \log \frac{e^{s(z^1_i, z_i^2)}}{e^{s(z_i^1, z_i^2)}+\sum_{j=1, j \neq i}^{N} e^{s(z_i^1, z_j^1)}}
\end{equation}
where $s$ denotes a cosine similarity function $s(z_i^1, z_i^2)=z_i^1\cdot z_i^2 /\|z_i^1\|\|z_i^2\|$, the superscript in $z_i^1$ and $z_i^2$ indicates the view index, $N$ indicates the number of training samples in a mini-batch. The model uncertainty constraint encourages the consistency between different views of semantic representation from the same input while enforcing the discrepancy between unrelated question pairs. By reducing the gap between the sub-models that contain different weight correspondences due to the randomness of the dropout mechanism, the robustness of the representation of the text-to-SQL parsers can be further enhanced. 




\subsection{\textbf{Uncertainty-aware Semantic Representation}}
We further augment the contextual representation $X_{I}$ of the input sample $I$ obtained by the encoder with the corresponding uncertainty-aware latent representation $z$ through the learnable gate $g$ as: 
\begin{equation}
    g=\operatorname{sigmoid}\left(z \cdot \mathbf{W}_{ z}+X_{I} \cdot \mathbf{W}_{x}\right)
\end{equation}
where $\mathbf{W}_{z}$, $\mathbf{W}_{x}$ denote the projection parameters.
Then, we formulate the overall semantic representation $\mathbf{U}\in \mathbb{R}^{\left|V^{n}\right| \times d}$ 
by combining the contextual representation $X_{I}$ and the uncertainty-aware latent representation $z$ with the corresponding gate $g$ as:
\begin{equation}
\mathbf{U}=\operatorname{LayerNorm}\left(g \cdot z+(1-g) \cdot X_{I}\right)\\
\end{equation}
where $\mathbf{U}$ is used as the input to the decoder (grammar-based decoder or T5 decoder) of the text-to-SQL parser to output the target SQL as $Y = \texttt{Decode}(\mathbf{U})$. We define the standard training objective of the encoder-decoder framework as $\mathcal{L}_{\rm T2S}$.

\paragraph{Joint Training} 
Finally, we combine the standard negative likelihood loss $\mathcal{L}_{\mathrm{T2S}}$ for SQL generation and the two uncertainty constraint loss functions ($\mathcal{L}_{\mathrm{DU}}$ and $\mathcal{L}_{\mathrm{MU}}$) to form the joint loss function $\mathcal{L}_{\text {total}}$ as follows:
\begin{equation}
\mathcal{L}_{\text {total }}
=\mathcal{L}_{\mathrm{T2S}}+\mathcal{L}_{\mathrm{DU}}+\mathcal{L}_{\mathrm{MU}}
\end{equation}



\begin{table}[t]  
    \centering
    \scalebox{0.9}{
    \begin{tabular}{lcc}  
    \toprule
    \textbf{Model}& ~~\textbf{EM}~~ & ~~~\textbf{EX}~~~ \\ 
    \midrule
    IRNet + BERT  & 61.9 & -  \\ 
    RAT-SQL + BERT  & 69.7 & - \\
    RAT-SQL + Grappa & 73.4 & - \\
    GAZP + BERT  & 59.1 &59.2\\ 
    BRIDGE + BERT  &65.5 &65.3\\ 
    BRIDGE v2 + BERT  &71.1 &70.3\\ 
    SMBOP + GRAPPA &74.7 &75.0\\ 
    RAT-SQL+GAP+NatSQL & 73.7 &75.0 \\ 
    \midrule
    LGESQL + ELECTRA  & 75.1 & - \\
    \rowcolor[RGB]{237,237,237} \quad w/ \sun & \textbf{76.8} & - \\
    PICARD + T5-Large  &69.1 & 72.9 \\
    \rowcolor[RGB]{237,237,237} \quad w/ \sun & {71.6} & \textbf{75.4}  \\
    \bottomrule
    \end{tabular}}
    \caption{Exact match ({EM}) and execution ({EX}) accuracy (\%) on \spider benchmark.}
    \label{tab:spider}
\end{table}

\begin{table}[t]
\centering
\scalebox{0.9}{
\begin{tabular}{lllc}
\toprule
{\multirow{2}*{\textbf{Model}}} & \multicolumn{2}{c}{~~\textbf{Dev.}~~} & ~~\textbf{Test}~~ \\
& {EM} & {EX}  & {EX} \\
    \midrule
    ETA + BERT & 50.10 & 68.30 & 54.10 \\
    \midrule
    ALIGN               &  43.70         &  62.10        & 50.10 \\
    \quad w/ BERT   &  48.40  &  67.70        & 54.30 \\
    \quad w/ RoBERTa      &  50.93         &  70.92        & 58.37 \\
    \rowcolor[RGB]{237,237,237} \quad\quad w/ \sun        & \textbf{52.93}  &\textbf{71.95}  & \textbf{59.34} \\      
    \bottomrule
\end{tabular}}
\caption{Exact match ({EM}) and execution ({EX}) accuracy (\%) on \squall dataset. }
\label{tab:squall}
\end{table}

\begin{table}[t]  
    \centering
    \scalebox{0.9}{
    \begin{tabular}{lccc}  
    \toprule
    \textbf{Model}& \textbf{\syn} & \textbf{\dk} & \textbf{\real}  \\ 
    \midrule
    GNN  & 23.6 & 26.0 & -\\ 
    IRNet & 28.4 & 33.1 & -\\ 
    RAT-SQL  & 33.6 & 35.8 & - \\
    RAT-SQL + BERT  & 48.2 & 40.9 & 58.1\\
    RAT-SQL + Grappa  & 49.1 &  38.5 &  59.3\\
    \midrule
    LGESQL + ELECTRA & 64.6 &  48.4 &  69.2\\
    \rowcolor[RGB]{237,237,237} \quad w/ \sun & \textbf{66.9} & \textbf{52.7} & \textbf{70.9}\\
    \bottomrule
    \end{tabular}
    }
    \caption{Exact match accuracy (\%) on \syn, \dk and  \real benchmark. }
    \label{tab:robust}
\end{table}

\begin{table*}[t]
\centering
\resizebox{0.9\hsize}{!}{
\begin{tabular}{lcccccccccc}
\toprule
\multirow{2}{*}{Model} & \multicolumn{5}{c}{\spider}             & \multicolumn{5}{c}{\syn}              \\
\cline{2-11}
        & easy & medium & hard & extra & all & easy & medium & hard & extra & all \\
\midrule
LGESQL+ELECTRA  &  91.9    &  78.3      &  64.9    &  \textbf{52.4}    & 75.1 &  79.4    &  67.9    & \textbf{62.1}     & 36.1      & 64.6   \\
\quad w/ \sun &\textbf{92.3}  &\textbf{80.3} &\textbf{70.7} &50.6 & \textbf{76.8}    & \textbf{79.8}    &\textbf{72.3 }    & 59.9     &  \textbf{41.4}     & \textbf{66.9}    \\
\midrule
\midrule
\multirow{2}{*}{Model} & \multicolumn{5}{c}{\dk}             & \multicolumn{5}{c}{\real}              \\
\cline{2-11}
        & easy & medium & hard & extra & all & easy & medium & hard & extra & all \\
\midrule
LGESQL+ELECTRA  & 74.5     & 46.7      & 41.9    & 29.5      &  48.4   & 86.2& \textbf{77.8}  &60.6  &41.2  &69.2        \\
\quad w/ \sun &\textbf{75.5}  &\textbf{53.7} &\textbf{47.3 } &\textbf{30.5} & \textbf{52.7}    & \textbf{89.9}    & 77.3   & \textbf{61.6 }     &  \textbf{45.4}      & \textbf{70.9}    \\
\bottomrule
\end{tabular}
}
\caption{Exact matching accuracy by varying the levels of difficulty of the inference data on the of \spider, \syn, \dk and \real.}
\label{diff_result}
\end{table*}

\begin{table*}[t]
    \centering
    \begin{tabular}{lcccc}
    \toprule
    \textbf{Model} & \textbf{\spider}   &  \textbf{\syn} & \textbf{\dk}   &  \textbf{\real} \\
    \midrule
    
     LGESQL+\sun & 76.8 & 66.9 & 52.7 & 70.9 \\
    \quad w/o $\mathcal{L}_{\rm DU}$  & 75.8 & 65.8 & 51.2 & 69.5 \\ 
    \quad w/o  $\mathcal{L}_{\rm MU}$  & 76.2 & 66.4 & 52.1 & 70.1 \\

    \bottomrule
    \end{tabular}
    \caption{Ablation results in terms of exact match accuracy  on \spider, \syn, \dk and \real.}
    \vspace{0.3cm}
    \label{tab:ablation}
\end{table*}




\section{Experimental Setup}
\subsection{Datasets}
We conduct extensive experiments on five benchmark datasets for text-to-SQL parsing. 
(1) \textbf{\spider} \cite{spider} is a large-scale cross-domain zero-shot text-to-SQL benchmark.
It originally contains 8659 training examples across 146 databases in total. 
We follow the common practice to report the exact match accuracy and execution accuracy.
(2) \textbf{\syn} \cite{spiderSYN} is a challenging variant of \spider, which consists of 1034 evaluation examples.
\syn is constructed by manually modifying NL questions in \spider using synonym substitution.
(3) \textbf{\dk} \cite{spiderDK} is constructed by selecting 535 samples from \spider dev set, where 270 pairs are the original \spider samples while the rest 265 pairs are modified by incorporating the domain knowledge. 
(4) \textbf{\real} \cite{deng2020structure} is a more realistic and challenging evaluation setting with explicit mentions of column names being manually removed. 
(5) \textbf{\squall} \cite{shi2020potential} is constructed by generating SQL queries of the English-language questions in \texttt{WIKITABLEQUESTIONS} \cite{pasupat2015compositional} and manually align questions with corresponding SQLs. It consists of 15,622 examples which are split into training (9,032), development (2,246) and test (4,344) sets. 
All datasets employed in this paper are in English.

\subsection{Baseline Methods}
We compare \sun with several strong baseline methods, including IRNet \cite{guo2019towards}, RAT-SQL \citep{ratsql}, GAZP \citep{zhong2020grounded}, BRIDGE \citep{lin2020bridging}, SMBOP \citep{rubin2020smbop}, LGESQL \cite{lgesql} and PICARD \cite{Scholak2021PICARDPI}. 
Since \sun is model-agnostic and potentially applicable for any neural text-to-SQL parsers, we adopt LGESQL and PICARD, which are the state-of-the-art graph-based and T5-based methods respectively, as our base models to verify the universality of \sun.
In addition, for \squall, we adopt the previous SOTA model ALIGN \cite{shi2020potential} with RoBERTa \cite{devlin2018bert} and ETA \cite{liu2021awakening} as our base model.

\subsection{Implementation Details}
For the LGESQL, following \cite{lgesql}, the hidden size of the graph attention network is set to 512 and the number of layers is set to 8. The number of heads in multi-head attention is 8 and the dropout rate is set to 0.2 for both the encoder and decoder. In the decoder, the dimension of hidden state, action embedding and node type embedding are set to 512, 128 and 128, respectively.
We use AdamW optimizer \cite{loshchilov2017decoupled} with linear warmup scheduler and the warmup ratio of total training steps is 0.1.
The learning rate is 1e-4 and the weight decay rate is 0.1. The batch size is set to be 20 and the maximum gradient norm is 5. 
The number of training epochs is 200.
For the PICARD \cite{Scholak2021PICARDPI}, we follow the official implementation to fine-tune T5-large for 400 epochs. We use Adafactor optimizer  \cite{shazeer2018adafactor} with a learning rate of 1e-4.
For the ALIGN model, we employ \sun on the RoBERTa setting. The representation size is 1024. The decoder is implemented with 2-layer LSTM and the hidden size is set to 128.  We adopt the Adam optimizer \cite{kingma2014adam} with learning rate of 0.001 and the dropout rate is set to be 0.3. 
Specifically, we employ kernels with window sizes ranging from 3 to 5 to obtain the global semantic representations used in the uncertainty measurement process in \sun.


\begin{table*}[t]  
\centering
\small
\scalebox{0.8}{
\begin{tabular}{llc}
\toprule
\rowcolor[RGB]{240,248,255} \textbf{\textcolor{black}{Case 1. (\spider)}} & & \\ 
 \midrule
  Gold & SELECT T1.template\_type\_code, count(*) FROM Templates AS T1 JOIN Documents AS T2 ON T1.template\_id & \sqlcorrect{} \\ 
 & = T2.template\_id GROUP BY T1.template\_type\_code &  \\
 \hdashline
 Question & \textit{Show all template type codes and the number of documents using each type.} &  \\
 LGESQL & SELECT T1.template\_type\_code, count(*) FROM Templates AS T1 JOIN Documents AS T2 ON T1.template\_id & \sqlcorrect{} \\
  & = T2.template\_id GROUP BY T1.template\_type\_code &  \\
 \quad w/ \sun& SELECT T1.template\_type\_code, count(*) FROM Templates AS T1 JOIN Documents AS T2 ON T1.template\_id & \sqlcorrect{} \\ 
 & = T2.template\_id GROUP BY T1.template\_type\_code &  \\
 \hdashline
 SE\_Question& \textit{What are the different template type codes, and how many documents use each type?} & \\
 LGESQL & SELECT Templates.Template\_Type\_Code , COUNT(*) FROM Templates GROUP BY Templates.Template\_Type\_Code & \sqlwrong{} \\
 \quad w/ \sun& SELECT T1.template\_type\_code, count(*) FROM Templates AS T1 JOIN Documents AS T2 ON T1.template\_id & \sqlcorrect{} \\ 
 & = T2.template\_id GROUP BY T1.template\_type\_code &  \\

 \midrule
 \rowcolor[RGB]{240,248,255} \textbf{\textcolor{black}{Case 2. (\dk)}} & & \\ 
  \midrule
 Gold & SELECT count(*) FROM student AS T1 JOIN has\_pet AS T2 ON T1.stuid  =  T2.stuid JOIN pets AS T3 ON T2.petid  =  T3.petid \\ &  WHERE T1.sex  =  'F' AND T3.pettype  =  'dog'
 \\ 
 \hdashline
 Question & \textit{How many puppy pets are raised by female students?} \\
 LGESQL & SELECT count(*) FROM student AS T1 JOIN has\_pet AS T2 ON T1.stuid  =  T2.stuid JOIN pets AS T3 ON T2.petid  =  T3.petid & \sqlcorrect{} \\ &  WHERE T1.sex  =  'F' AND T3.pettype  =  'dog'\\
 \quad w/ \sun& SELECT count(*) FROM student AS T1 JOIN has\_pet AS T2 ON T1.stuid  =  T2.stuid JOIN pets AS T3 ON T2.petid  =  T3.petid & \sqlcorrect{} \\ &  WHERE T1.sex  =  'F' AND T3.pettype  =  'dog'\\
 \hdashline
 SE\_Question& \textit{Find the number of puppy pets that are raised by female students (with sex F).}\\
 LGESQL & SELECT COUNT(*) FROM Pets JOIN Has\_Pet JOIN Student WHERE Student.Sex = 'F'  & \sqlwrong{} \\
 \quad w/ \sun& SELECT count(*) FROM student AS T1 JOIN has\_pet AS T2 ON T1.stuid  =  T2.stuid JOIN pets AS T3 ON T2.petid  =  T3.petid & \sqlcorrect{} \\ &  WHERE T1.sex  =  'F' AND T3.pettype  =  'dog'\\

 \bottomrule
\end{tabular}}
\caption{Case study: cases are sampled from \spider, and \dk. SE indicates semantically-equivalent. }
\vspace{0.5cm}
\label{tab:case}
\end{table*}

\section{Experimental Results}
\subsection{Main Results}
\paragraph{Results on \spider}
Table \ref{tab:spider} shows the exact match accuracy (EM) and execution accuracy (EX) scores of our method and compared baselines on the \spider dataset. We observe that  \sun can bring substantial improvements, which achieves a notable gain of 1.7\% on the exact match accuracy scores over the strongest baseline LGESQL. In addition,  PICARD+\sun obtains significantly better results than the compared baseline methods on \spider dataset in terms of exact match accuracy and execution accuracy, which both achieve superior results of 2.5\% improvement. 

\paragraph{Results on more challenging and realistic settings}
To evaluate the effectiveness of \sun in more challenging and realistic settings, Table \ref{tab:robust} illustrates the experimental results on \syn, \dk and \real datasets. We can observe that \sun consistently and substantially surpasses the compared models by a noticeable margin on three datasets in terms of exact match accuracy. 
In particular, \sun  achieves considerable improvement over LGESQL on all three datasets. For \syn,  LGESQL with \sun outperforms LGESQL by 2.3\% EM score, demonstrating that \sun can improve the robustness of text-to-SQL parsers to synonym substitution. An improvement of 4.3\% EM score is observed on \dk benchmark, verifying that \sun contributes to  the generalization ability of text-to-SQL parsers to unseen domains. Furthermore, under the  challenging \real setting where all explicit mentions of column names are removed, LGESQL+\sun also achieves a strong performance (70.9\%) which is 1.7\% 
higher than LGESQL.


\paragraph{Results on \squall}
Table \ref{tab:squall} shows the experimental results on \squall. Since \squall does not contain target SQL queries in the test set, we merely present the execution accuracy score on test set. Overall, \sun significantly improves the performance on both the dev and test sets of \squall, achieving the gains of 2.0\% EM score and 1.03\% EX score on dev set, and 1.97\% EX score on test set over the strongest baseline ALIGN+RoBERTa on \squall. 

\subsection{Results on Complex Queries}
We investigate the performance of \sun on different queries. The SQL queries in the \spider, \syn, \dk and \real benchmarks are divided into four levels (i.e., easy, medium, hard, extra hard) based on their difficulty, where the difficulty is defined
based on the number of SQL components. Table \ref{diff_result}  summarizes the results of the four benchmarks with four levels of difficulty. From the results, we can observe that \sun can boost the performance of LGESQL across almost all different difficulty levels on the four datasets. In particular, LGESQL+\sun obtains much better performance on the extremely hard samples than LGESQL by a large margin. For example, LGESQL with \sun shows 4.7\%, 1.0\% and 4.2\% improvements on the extra hard samples in \syn, \dk and \real respectively.


\subsection{Ablation Study}
To analyze the impact of two kinds of uncertainties in \sun, we also conduct an ablation test by discarding the data uncertainty constraint (denoted as w/o $\mathcal{L}_{\rm DU}$) and the model uncertainty constraint (denoted as w/o $\mathcal{L}_{\rm MU}$), respectively. The ablation results are summarized in Table \ref{tab:ablation}. As expected, both uncertainty constraints contribute great improvements to \sun.
For example, the performance of \sun w/o $\mathcal{L}_{\rm DU}$ decreases by 1.1/1.0/1.5 points on the \syn/\spider/\dk dev sets, verifying that the data uncertainty constraint is essential for improving the model performance. 
In addition, the performance of \sun decreases by 0.8/0.5 points on \real/\syn evaluation sets when removing the model uncertainty constraint.

\subsection{Case Study}
In this section, we present two cases sampled from \spider and \dk to demonstrate the effectiveness of \sun qualitatively. As illustrated in Table \ref{tab:case}, we report the original questions, the semantically-equivalent variant of the original questions (SE-question), the gold SQL queries, and the SQL queries generated by LGESQL and LGESQL+\sun. From the results, we can observe that LGESQL+\sun can generate more accurate SQL queries than LGESQL. Taking the first case as an example, although both LGESQL and  LGESQL+\sun can generate accurate SQL queries given the question, LGESQL fails to correctly understand its semantically-equivalent question and thus generates inappropriate SQL where the necessary mention of database schema ``\textit{Document}'' is ignored. In contrast, LGESQL+\sun generates the same correct SQL for both the original question and its semantically-equivalent question by improving the robustness of the text-to-SQL parsers with data and model uncertainty constraints. We observe similar trends on the \dk dataset. Concretely, LGESQL+\sun can generate consistent SQL for two semantically-equivalent questions that have very different expressions, while LGESQL cannot handle such cases. 

The effectiveness of \sun according to the observed results illustrated in Table \ref{tab:case} demonstrates that the exploration of underlying complementary semantic information in \sun helps text-to-SQL parsers understand the different variants of expressions and generate better SQL queries with comprehensive semantics.


\section{Conclusion}
In this paper, we proposed a novel \sun method to explore intrinsic uncertainties in text-to-SQL parsing. 
First, we devised a data uncertainty constraint to capture complementary semantic information among multiple semantically-equivalent questions and thus improve the robustness of the text-to-SQL parsers. 
Second, a model uncertainty constraint was leveraged to refine the representations by encouraging high similarity between the output representations of two perturbed encoding networks for the same input question.
Experimental results on five benchmark datasets showed that \sun significantly outperformed the compared methods.

\section{Acknowledges}
My acknowledges: This work was partially supported by National Natural Science Foundation of China (No. 61906185), Youth Innovation Promotion Association of CAS China (No. 2020357), Shenzhen Science and Technology Innovation Program (Grant No. KQTD20190929172835662), Shenzhen Basic Research Foundation (No. JCYJ20210324115614039 and No. JCYJ20200109113441941). This work was supported by Alibaba Group through Alibaba Innovative Research Program.

\bibliography{anthology,custom}




\end{document}